\documentclass[sigconf]{acmart}

\AtBeginDocument{%
  }

\setcopyright{none}
\renewcommand\footnotetextcopyrightpermission[1]{}
\acmConference[KDD 2026 Workshop GMLLM]{The 2nd Frontiers in Graph Machine Learning for the Large Model Era}{August 2026}{Jeju, Republic of Korea}

\author{Woohyun Lee}
\affiliation{%
\institution{Sungkyunkwan University}
\city{Suwon}
\country{Republic of Korea}
}

\author{Hogun Park}
\affiliation{%
\institution{Sungkyunkwan University}
\city{Suwon}
\country{Republic of Korea}
}

\usepackage{booktabs}
\usepackage{multirow}
\usepackage{amsmath}
\usepackage{amsfonts}
\usepackage{balance}

\begin{document}

\title{CondPSE: A Polynomial-Filtered Structural Encoder with Conditional Modulation for Graphs}

\begin{abstract}
\emergencystretch=1em
\textcolor{black}{Message-passing graph neural networks are bounded by the 1-WL test and can miss topological structure that distinguishes non-isomorphic graphs. Positional and structural encodings (PSE) inject such topology-derived signals, and learned PSE encoders such as GPSE pretrain a single encoder to produce these signals from random node probes, which can then be frozen and reused as inputs across downstream graph models. We present CondPSE, a learned PSE encoder that applies a learnable polynomial graph filter bank to standard Gaussian node probes and refines the resulting structural-response branches through FiLM-style modulation conditioned on cross-filter, local message-passing, and graph-level signals. CondPSE is pretrained to reconstruct node-level positional/structural targets and graph-level invariants, and is then frozen for use as a downstream input encoding. On synthetic structural-discrimination benchmarks, CondPSE separates graph structures that 1-WL-bounded message passing cannot: it raises CSL accuracy from 42.9\% to 97.3\% and EXP accuracy from 68.3\% to 99.9\% relative to GPSE, and ablations show that the polynomial filter bank accounts for most of this gain. On real molecular property prediction, the picture is more limited. With a hybrid local-message-passing/global-attention backbone, CondPSE performs comparably to GPSE without surpassing it, and a ZINC backbone sweep shows no consistent ordering between the two encoders. We report these results and discuss why strong synthetic structural
discrimination does not, on its own, yield a downstream advantage
for frozen learned PSE encoders, including the role of downstream
integration and possible mismatch between structural pretraining
targets and molecular property labels.}
\end{abstract}



\maketitle

\section{Introduction}

Message-passing GNNs are bounded by the 1-WL test~\cite{morris2019weisfeiler,xu2019how}. Structural and positional encodings add topology-derived signals that such models may miss~\cite{dwivedi2022graph,kreuzer2021rethinking,NEURIPS2021_f1c15925,rampasek2022recipe,pmlr-v235-canturk24a}. Hand-crafted encodings include random-walk statistics~\cite{rampasek2022recipe}, Laplacian eigenvectors~\cite{dwivedi2022graph}, shortest-path distances, and centrality encodings~\cite{NEURIPS2021_f1c15925}; learned PSE encoders such as GPSE~\cite{pmlr-v235-canturk24a} are instead pretrained to produce positional/structural encodings by reconstructing multiple positional/structural targets. Because pretrained PSE encoders can be frozen and used to generate positional/structural inputs for downstream graph models, their value depends not only on what positional/structural targets they reconstruct, but also on how downstream architectures use the resulting encodings.

We study this gap through \emph{CondPSE}, a learned structural encoder that applies polynomial graph filters to standard Gaussian node probes. CondPSE is pretrained to reconstruct node-level positional/structural targets and graph-level invariants. Relative to GPSE's~\cite{pmlr-v235-canturk24a} stacked message-passing extractor, CondPSE uses a polynomial graph filter bank to initialize multiple structural-response branches before conditional modulation. It modulates these branches using cross-filter, local, and global context via FiLM-style modulation~\cite{perez2018film}.

Our main finding is that stronger benchmark-level structural discrimination obtained using frozen node encodings does not automatically translate into stronger downstream graph-level transfer. CondPSE achieves strong performance on CSL and near-perfect performance on EXP, suggesting that the encoder separates structural patterns useful for CSL and EXP, where labels are determined by graph topology rather than node or edge attributes. In our molecular evaluation, transfer with the hybrid local-message-passing/global-attention backbone remains close to GPSE, and the ZINC backbone sweep shows that the relative performance of GPSE and CondPSE varies by downstream architecture despite CondPSE's stronger CSL/EXP structural-discrimination performance.

We use CondPSE as a diagnostic encoder for studying when benchmark-level structural discrimination transfers to downstream graph-level prediction.

We make three contributions:
\begin{itemize}
    \item We introduce CondPSE, a learned structural encoder that uses polynomial-filtered structural-response branches followed by conditional modulation.
    \item We find that CondPSE obtains strong CSL/EXP performance, and ablations indicate that the polynomial filter bank plays the largest role while cross-filter, local, and global conditioning provide additional contributions.
    \item We show that stronger CSL/EXP structural-discrimination performance does not automatically translate into stronger downstream molecular transfer, and use a ZINC backbone sweep to examine how downstream architecture affects the relative utility of GPSE and CondPSE.
\end{itemize}

\section{Related Work}

\paragraph{Structural and Positional Encodings.}
Many graph learning models augment node features with positional or structural information to overcome message-passing limits. Traditional message-passing graph neural networks (MPNNs) such as GCN~\cite{kipf2017semi}, GIN~\cite{xu2019how}, and GINE~\cite{hu2020pretraining} are bounded by the 1-WL graph isomorphism test~\cite{xu2019how,morris2019weisfeiler}. Traditional positional/structural encodings include Laplacian eigenvectors (LapPE) and random-walk statistics (RWSE). To integrate these features, LSPE~\cite{dwivedi2022graph} learns decoupled structural and positional representations, while SAN~\cite{kreuzer2021rethinking} and Graphormer~\cite{NEURIPS2021_f1c15925} incorporate them into Transformer architectures. GraphGPS~\cite{rampasek2022recipe} combines local message passing, global attention, and positional encodings in a modular graph Transformer backbone.

\paragraph{Polynomial Graph Filtering and Spectral GNNs.}
Spectral graph neural networks define graph filters through the eigenstructure of graph operators, while practical models often implement such filters using polynomial functions of the adjacency or Laplacian matrix. ChebNet~\cite{defferrard2016convolutional} approximates spectral graph filters with Chebyshev polynomials, replacing explicit eigendecomposition with polynomial functions of a graph operator. GPR-GNN~\cite{chien2021adaptive} learns weights over multiple propagation powers. These methods motivate using powers of a graph operator to capture responses at different propagation depths. CondPSE uses polynomial filtering as an extraction stage: it forms initial structural-response branches from standard Gaussian node probes, then refines these branches through cross-filter, local, and global conditioning before producing the final structural encoding.

\paragraph{Random Features and Learned Encoders.}
Random node features have been shown to increase the distinguishing power of GNNs. Sato et al.~\cite{sato2021random} and Abboud et al.~\cite{abboud2021surprising} show that random node initialization can increase the distinguishing power of GNNs beyond deterministic message passing in certain settings. GPSE~\cite{pmlr-v235-canturk24a} builds on this by learning a transferable positional and structural encoder from random features, using a GatedGCN~\cite{bresson2017residual} backbone to reconstruct multiple hand-crafted PSE targets. Unlike fixed encodings such as LapPE or RWSE, learned random-feature encoders do not directly prescribe a single structural statistic; instead, they learn to convert graph-induced random responses into positional/structural encodings. CondPSE shares GPSE's~\cite{pmlr-v235-canturk24a} goal of pretraining a transferable PSE encoder but constructs its intermediate responses differently: it first produces polynomial-filtered probe responses and then adapts them with conditional modulation modules, including local message passing.


\section{Methodology}
\label{sec:methodology}

CondPSE starts from NormalSE probes sampled i.i.d. from a standard Gaussian distribution. These probes are graph-agnostic before adjacency propagation. The polynomial filter bank applies powers of the adjacency matrix to the probes and forms $K$ learned mixtures of propagation-depth responses. These mixtures serve as initial structural-response branches that are later refined by conditioning modules. The conditioning modules use cross-filter comparisons, local message passing, and graph-level statistics, and FiLM modulation applies the resulting conditioning signals as adaptive channel-wise transformations. The encoder is pretrained once and then frozen; downstream experiments therefore evaluate how fixed structural encodings are incorporated by each backbone.

\subsection{Extraction Phase: Polynomial Graph Filter Bank}

Given a graph $G=(\mathcal{V}, \mathcal{E})$ with node set $\mathcal{V}$, edge set $\mathcal{E}$, and unnormalized adjacency matrix $A$, we sample stochastic node probes $Z \in \mathbb{R}^{N \times d_{\mathrm{in}}}$ as $Z_{v,c}\sim\mathcal{N}(0,1)$ and compute propagation responses from NormalSE probes. The probes serve two roles. First, stochastic node probes break node symmetry and can help distinguish structures that deterministic 1-WL-style message passing may fail to separate~\cite{abboud2021surprising,sato2021random}. Second, they act as random inputs for querying a matrix operator, analogous to randomized numerical linear algebra methods that use random vectors to estimate properties of matrix functions and operators~\cite{hutchinson1990stochastic,ubaru2017fast}. In CondPSE, $A^0Z$ preserves the unpropagated probe channel, while $A^iZ$ for $i\geq 1$ records the same probes after $i$ adjacency propagations. Thus, graph-structural information is introduced through adjacency propagation rather than through the initial probe values. Each polynomial filter learns a mixture of these propagation-depth responses, and the resulting filter branches serve as initial structural-response branches for the conditioning modules.

Following polynomial graph filtering ideas used in spectral GNNs such as ChebNet~\cite{defferrard2016convolutional} and GPR-GNN~\cite{chien2021adaptive}, we filter $Z$ through a bank of $K$ learnable polynomial graph filters of degree $d$:
\begin{equation}
\tilde{H}_k = \sum_{i=0}^d \bar{c}_{k,i} A^i Z, \quad H_k^{(0)} = \mathrm{LN}(\tilde{H}_k W_{\text{proj}} + b_{\text{proj}}),
\end{equation}
where $\bar{c}_{k,i}$ denotes the normalized polynomial coefficient:
\begin{equation}
\bar{c}_{k,i} = \frac{c_{k,i}}{\sqrt{\sum_{j=0}^d c_{k,j}^2} + \epsilon},
\end{equation}
and $W_{\text{proj}} \in \mathbb{R}^{d_{\text{in}} \times D}$ projects features into the model dimension $D$. Here and below, $\mathrm{LN}$ denotes LayerNorm. We use coefficient normalization and LayerNorm after projection to keep propagation responses numerically stable across graphs, filters, and propagation degrees. A single filter provides one learned mixture of propagation responses, while the polynomial graph filter bank computes multiple such responses across filters. Thus, each branch is a learned polynomial filter over $A$, implemented through repeated adjacency applications rather than eigendecomposition. The $K$ branches retain different mixtures of propagation depths before conditioning. After $B$ conditioning blocks, the $K$ filter branches are fused into a single node encoding:
\begin{equation}
E_v = \psi\!\left(H^{(B)}_{v,1} \mathbin{\Vert} \cdots \mathbin{\Vert} H^{(B)}_{v,K}\right),
\qquad
\psi: \mathbb{R}^{KD} \to \mathbb{R}^{d_{\mathrm{out}}}.
\end{equation}
where $\mathbin{\Vert}$ denotes feature concatenation, $d_{\mathrm{out}}$ is the output node-encoding dimension, and $\psi$ is a two-layer MLP that produces the final node encoding~$E_v$.

\subsection{Conditioning Modulation and Injection}

At each block $b \in \{1, \dots, B\}$, the hidden tensor $H^{(b-1)} \in \mathbb{R}^{N \times K \times D}$ is modulated using three conditioning signals derived from the current filter-branch states:

\textbf{Cross-filter conditioning:} Cross-filter conditioning computes pairwise comparison features across polynomial-filter branches, allowing information exchange among current filter-branch states. For each node $v$ and filter pair $(k, j)$, we compute comparison features from the current hidden tensor:
\[
F_{v,k,j} = \bigl[ H^{(b-1)}_{v,k} \mathbin{\Vert} H^{(b-1)}_{v,j} \mathbin{\Vert} (H^{(b-1)}_{v,k} - H^{(b-1)}_{v,j}) \mathbin{\Vert} (H^{(b-1)}_{v,k} \odot H^{(b-1)}_{v,j}) \bigr].
\]
Pairwise messages are summed over $j \ne k$ and mapped through shared MLPs to yield the cross-filter signal $I_{\text{xf}, v, k} \in \mathbb{R}^D$.

\textbf{Local conditioning:} Local conditioning applies a shared two-layer ResGatedGCN~\cite{bresson2017residual} over each filter branch independently, aggregating edge-neighborhood information within each branch to produce the local spatial signal $I_{\text{loc}, v, k} \in \mathbb{R}^D$.

\textbf{Global conditioning:} Global conditioning computes graph-wide statistics, including mean, standard deviation, and maximum across nodes for each filter branch, projects them through an MLP, and broadcasts the resulting signal to nodes as $I_{\text{glb}, v, k} \in \mathbb{R}^D$.

\begin{sloppypar}
These signals are concatenated into $U_{v, k} = [ I_{\text{xf}, v, k} \mathbin{\Vert} I_{\text{loc}, v, k} \mathbin{\Vert} I_{\text{glb}, v, k} ] \in \mathbb{R}^{3D}$. The block-specific FiLM MLP maps $U_{v,k}$ to a raw scale vector and shift vector, $(\gamma^r_{v,k}, \beta_{v,k}) = \phi_b(U_{v,k})$, where $\gamma^r_{v,k}, \beta_{v,k} \in \mathbb{R}^{D}$. The final channel-wise scale vector is constrained around one by $\gamma_{v,k} = \mathbf{1} + \tanh(\gamma^r_{v,k})$:
\end{sloppypar}
\begin{equation}
\begin{aligned}
(\gamma^r_{v,k}, \beta_{v,k}) &= \phi_b(U_{v,k}), \quad
\gamma_{v,k}=\mathbf{1}+\tanh(\gamma^r_{v,k}),\\
H_{v,k}^{(b)} &= \mathrm{LN}\!\left(H_{v,k}^{(b-1)}+
s_{\mathrm{res}}(\gamma_{v,k}\odot H_{v,k}^{(b-1)}+\beta_{v,k})\right).
\end{aligned}
\end{equation}
Here, $\phi_b$ is the block-specific FiLM MLP, and $s_{\text{res}} \in \mathbb{R}$ is a learnable residual scale initialized to \(1\). Thus, each filter branch receives a learned channel-wise scale and shift.

\subsection{Downstream Integration}

Once pretrained, the CondPSE encoder is frozen. The node-encoding matrix $E \in \mathbb{R}^{N \times d_{\text{out}}}$, whose rows are $E_v$, is passed through a lightweight downstream adapter $f_{\text{PE}}$ (a linear or MLP projection) to produce $E' \in \mathbb{R}^{N \times d_{\text{pe}}}$. The final input to the downstream GNN is formed by concatenating the projected raw node features with the structural encoding:
\begin{equation}
X_{\text{in}} = \bigl[ X_{\text{raw}} W_x \mathbin{\Vert} E' \bigr] ,
\end{equation}
where $W_x \in \mathbb{R}^{d_{\text{in}} \times (d_{\text{emb}} - d_{\text{pe}})}$ projects the raw features to a complementary dimension before concatenation with $E'$. Freezing the encoder helps isolate the effect of the learned structural encoding under a fixed downstream integration setting.

\section{Experiments}
\label{sec:experiments}

We pretrain CondPSE on OGBG-MolPCBA~\cite{hu2020open}
using the same positional/structural pretraining targets as
GPSE~\cite{pmlr-v235-canturk24a} and evaluate the frozen CondPSE
encoder (\(B=6\) blocks, \(K=4\) filters, and degree \(d=15\)) following GPSE's public evaluation setting, including dataset splits, seed protocols, and downstream backbone configurations. Pretraining uses
graph-structure-derived targets and no downstream property labels.
We report molecular transfer results using a GraphGPS (GPS) backbone and, for ZINC, an additional downstream backbone sweep across GCN~\cite{kipf2017semi}, GIN~\cite{xu2019how},
GatedGCN~\cite{bresson2017residual}, GINE~\cite{hu2020pretraining},
and a Transformer-style backbone. Encoder, pretraining, downstream
training details, baseline sources, and supervised objectives are given in
Appendices~\ref{sec:appendix-encoder}--\ref{sec:appendix-downstream}.

\subsection{Benchmark-Level Structural Discrimination}

We evaluate benchmark-level structural discrimination through GIN-based performance on the Circular Skip Link (CSL)~\cite{murphy2019relational} and EXP~\cite{abboud2021surprising} graph isomorphism benchmarks. The CSL dataset consists of skip-link graphs where standard 1-WL GNNs struggle to distinguish different link lengths due to uniform degree distributions. The EXP dataset is a synthetic graph isomorphism benchmark designed to test GNN expressiveness against 1-WL limits.

\begin{sloppypar}
Following Cant\"{u}rk et al.~\cite{pmlr-v235-canturk24a}, we compare CondPSE with GPSE and standard positional/structural encodings. LapPE uses the first four non-trivial Laplacian eigenvectors, while RWSE uses 20-dimensional random-walk return probabilities. Table~\ref{tab:expressivity} reports the results.
\end{sloppypar}

\begingroup
\setlength{\intextsep}{0.4\baselineskip}
\begin{table}[H]
\centering
\caption{Synthetic structural-discrimination benchmarks using ten-times stratified five-fold cross-validation (ACC \% $\uparrow$). Bold indicates the best result for each benchmark.}
\label{tab:expressivity}
\footnotesize
\begin{tabular}{lcc}
\toprule
\textbf{Model} & \textbf{CSL} & \textbf{EXP} \\
\midrule
GIN (No PE) & $10.0 \pm 0.0$ & $48.7 \pm 2.2$ \\
GIN + GPSE & $42.9 \pm 7.9$ & $68.3 \pm 7.5$ \\
GIN + LapPE & $92.5 \pm 4.2$ & $99.5 \pm 0.8$ \\
GIN + RWSE & $\mathbf{100.0 \pm 0.0}$ & $99.7 \pm 0.6$ \\
\midrule
GIN + CondPSE w/o filter bank & $12.0 \pm 3.4$ & $23.6 \pm 2.2$ \\
GIN + CondPSE w/o global \textcolor{black}{conditioning}& $87.3 \pm 2.5$ & $99.5 \pm 0.3$ \\
GIN + CondPSE w/o local \textcolor{black}{conditioning}& $92.0 \pm 5.0$ & $99.5 \pm 0.3$ \\
GIN + CondPSE w/o \textcolor{black}{cross-filter conditioning} & $96.4 \pm 1.6$ & $99.8 \pm 0.5$ \\
GIN + CondPSE & $97.3 \pm 2.5$ & $\mathbf{99.9 \pm 0.2}$ \\
\bottomrule
\end{tabular}
\end{table}
\endgroup

\begin{sloppypar}
As shown in Table~\ref{tab:expressivity}, GIN without positional/structural encodings performs poorly on CSL and EXP, while explicit graph-operator encodings such as LapPE and RWSE provide strong positional/structural baselines. RWSE attains the highest CSL accuracy, which is consistent with the nature of CSL: the classes differ by circular skip-link structure under uniform degree patterns, and multi-step random-walk statistics can differentiate such structural differences. CondPSE reaches comparable CSL accuracy without concatenating RWSE directly as the downstream input encoding, although RWSE is included among the positional/structural pretraining targets. Thus, the comparison tests whether positional/structural supervision can be captured by a frozen learned encoder, rather than whether RWSE is explicitly supplied to the downstream GIN. Compared with GPSE, the closest learned random-feature encoder baseline, CondPSE increases CSL accuracy from 42.9 to 97.3 and EXP accuracy from 68.3 to 99.9. Compared with GPSE's stacked message-passing extractor, CondPSE explicitly forms multiple propagation-depth responses before conditional modulation. This design may make CSL/EXP-relevant structural differences easier for the downstream GIN to access. The ablations further separate
the roles of the extraction and conditioning stages: removing the
polynomial filter bank causes the largest degradation, suggesting that
propagation-depth-specific graph-operator responses provide important initial
inputs for the conditioning pipeline. Removing global, local, or
cross-filter conditioning yields smaller but visible drops, especially on
CSL, indicating that these contexts refine the filtered responses rather
than replace the extraction stage. Together, these results suggest that
CondPSE benefits from both polynomial-filtered initialization and
subsequent conditional modulation.
\end{sloppypar}

\subsection{GPS Molecular Downstream Transfer}

We evaluate molecular downstream transfer with a GraphGPS (GPS) backbone, which combines local message passing with global attention~\cite{rampasek2022recipe}. We test on ZINC (subset)~\cite{dwivedi2023benchmarking}, MolHIV and MolPCBA~\cite{hu2020open}, and PCQM4Mv2-subset~\cite{hu2021ogblsc}, using the metrics shown in Table~\ref{tab:downstream}.

\begingroup
\setlength{\intextsep}{0.2\baselineskip}
\setlength{\abovecaptionskip}{0pt}
\setlength{\belowcaptionskip}{2pt}
\renewcommand{\arraystretch}{0.82}
\begin{table}[h]
\centering
\caption{GPS molecular downstream prediction results averaged over 10 seeds. Bold indicates the best result for each dataset.}
\label{tab:downstream}
\scriptsize
\resizebox{\columnwidth}{!}{%
\begin{tabular}{lcccc}
\toprule
\textbf{Dataset} & \textbf{Metric} & \textbf{GPS + None} & \textbf{GPS + GPSE} & \textbf{GPS + CondPSE} \\
\midrule
ZINC (subset) & MAE $\downarrow$ & $0.1182 \pm 0.0049$ & $\mathbf{0.0648 \pm 0.0030}$ & $0.0649 \pm 0.0035$ \\
MolHIV & AUROC $\uparrow$ & $0.7798 \pm 0.0077$ & $\mathbf{0.7815 \pm 0.0133}$ & $0.7797 \pm 0.0070$ \\
MolPCBA & AP $\uparrow$ & $0.2869 \pm 0.0012$ & $0.2911 \pm 0.0036$ & $\mathbf{0.2922 \pm 0.0033}$ \\
PCQM4Mv2-subset & MAE $\downarrow$ & $0.1329 \pm 0.0030$ & $\mathbf{0.1196 \pm 0.0004}$ & $0.1207 \pm 0.0006$ \\
\bottomrule
\end{tabular}
}
\end{table}
\endgroup

CondPSE is close to GPSE across the molecular tasks: it is slightly better on MolPCBA and slightly worse on ZINC, MolHIV, and PCQM4Mv2-subset. These small differences motivate a controlled ZINC sweep over downstream backbones.

\subsection{ZINC Downstream Backbone Sweep}

To test whether the relative utility of learned structural encodings varies across downstream architectures, we run a ZINC downstream backbone sweep~\cite{dwivedi2023benchmarking}. For each architecture, we compare no structural encoding with encodings from GPSE and CondPSE under the same ZINC setup. Table~\ref{tab:backbone} reports ZINC MAE for each backbone and encoding choice.

\begin{table}[H]
\centering
\caption{ZINC downstream backbone sweep averaged over 4 seeds (MAE $\downarrow$). Bold indicates the best result for each architecture.}
\vspace{-0.15in}
\label{tab:backbone}
\footnotesize
\begin{tabular}{lccc}
\toprule
\textbf{Backbone} & \textbf{None} & \textbf{GPSE} & \textbf{CondPSE} \\
\midrule
GCN & $0.288 \pm 0.004$ & $0.129 \pm 0.003$ & $\mathbf{0.126 \pm 0.003}$ \\
GatedGCN & $0.236 \pm 0.008$ & $0.113 \pm 0.003$ & $\mathbf{0.110 \pm 0.002}$ \\
GIN & $0.285 \pm 0.004$ & $\mathbf{0.124 \pm 0.002}$ & $0.131 \pm 0.004$ \\
GINE & $0.118 \pm 0.005$ & $\mathbf{0.065 \pm 0.003}$ & $0.069 \pm 0.002$ \\
Transformer & $0.686 \pm 0.017$ & $0.189 \pm 0.016$ & $\mathbf{0.186 \pm 0.014}$ \\
\bottomrule
\end{tabular}
\end{table}
\vspace{-0.15in}
Both learned encodings improve over the no-encoding baseline for every tested backbone, suggesting that these structural encodings provide useful information in this ZINC evaluation. However, the relative performance is mixed: CondPSE has lower MAE than GPSE with GCN~\cite{kipf2017semi}, GatedGCN~\cite{bresson2017residual}, and the Transformer-style backbone~\cite{NEURIPS2021_f1c15925,rampasek2022recipe}, but higher MAE with GIN~\cite{xu2019how} and GINE~\cite{hu2020pretraining}. Because several gaps are within reported variation, Table~\ref{tab:backbone} is best read as a backbone-sensitivity check rather than a definitive superiority claim. This mixed pattern indicates that CSL/EXP performance alone does not determine downstream transfer.

\section{Discussion}
\label{sec:discussion}

The results separate benchmark-level structural discrimination from downstream utility. We discuss this gap through backbone-dependent transfer, target alignment, and generalization.

\paragraph{Backbone-Dependent Transfer.} Table~\ref{tab:backbone} suggests that transfer behavior varies by architecture, although many GPSE--CondPSE differences are small. Both learned encodings improve over the no-encoding baseline across the ZINC backbones, but the mixed GPSE--CondPSE results suggest that their utility is not determined by CSL/EXP performance alone. We view the sweep as a diagnostic check against the hypothesis that stronger CSL/EXP structural-discrimination performance should consistently produce better downstream transfer. These results do not identify a single architectural factor that explains all trends, but they suggest that downstream integration is a relevant factor: different backbones may combine raw features, message passing, and positional/structural information in ways that favor different structural encoders.

\paragraph{Target Alignment Considerations.} Beyond downstream integration, the choice of pretraining targets affects transfer, potentially leading to negative transfer~\cite{hu2020pretraining}. Reconstructing graph-level invariants or hand-crafted positional/structural targets from node-level encodings may introduce a possible target-alignment mismatch. CondPSE reconstructs node-level positional/structural targets (ElectrostaticPE, LapPE, RWSE, HKdiagSE) and graph-level invariants (EigVals, CycleGE). These targets may not align perfectly with molecular labels, so optimizing these reconstruction objectives or achieving strong benchmark-level structural discrimination may not guarantee a task-aligned downstream inductive bias. We do not directly measure target-label alignment, so this remains a hypothesis. Future research should evaluate objectives that more directly anticipate downstream label structure.

\paragraph{Expressivity-Generalization Trade-off.} The gap between strong CSL / near-perfect EXP performance and molecular transfer close to GPSE highlights a potential generalization trade-off. This empirical pattern is consistent with recent theory suggesting that greater expressivity can hurt generalization when structural distinctions reduce intra-class concentration or are misaligned with the downstream task metric~\cite{maskey2026graph,li2025towards}. One possible explanation, not directly tested here, is that highly expressive structural encodings may induce overly fine structural partitions, making it harder for the downstream model to ignore irrelevant topological variation. Practical datasets like ZINC and MolHIV require models to generalize across varied graphs with similar properties. If structural encodings vary substantially for graphs with similar molecular properties, the downstream model may struggle to generalize.

\paragraph{Scope and Future Directions.} The present experiments diagnose transfer behavior for learned structural encoders across the evaluated downstream architectures. Future work should isolate how cross-filter modulation, local message passing, and global context contribute under different graph families; extend the downstream backbone sweep beyond ZINC; and study positional/structural objectives that better anticipate task-specific semantics. More broadly, the pretraining objective, encoder design, and downstream adapter should be studied as interacting factors in transfer.

\section{Conclusion}
\label{sec:conclusion}

\begin{sloppypar}
CondPSE serves as a diagnostic encoder for studying the alignment between benchmark-level structural discrimination and downstream transfer. It achieves strong CSL/EXP performance, while molecular transfer remains close to GPSE and the ZINC sweep shows architecture-dependent relative performance. For graph pipelines that use frozen learned PSE encoders as inputs to downstream backbones, these mixed transfer results suggest that benchmark-level structural discrimination alone does not by itself determine downstream transfer behavior.
\end{sloppypar}

\clearpage
\balance
\section*{Acknowledgements}
This work was supported by the Institute of Information \& Communications Technology Planning \& Evaluation (IITP) and the National Research Foundation of Korea (NRF), both funded by the Ministry of Science and ICT (MSIT), under Grant Nos. RS-2025-24803185, RS-2019-II190421, and IITP-2025-RS-2020-II201821.
\bibliographystyle{ACM-Reference-Format}
\setlength{\bibsep}{0.0pt}
\bibliography{main}

\clearpage
\nobalance
\appendix
\balance

\section{Implementation and Training Details}
\label{sec:appendix-details}

\subsection{CondPSE Encoder Configuration}
\label{sec:appendix-encoder}

We use a standard CondPSE encoder configuration for all reported experiments. The architectural specifications are summarized in Table~\ref{tab:condpse_config}.

\begin{table}[H]
\centering
\caption{CondPSE encoder configuration.}
\label{tab:condpse_config}
\footnotesize
\begin{tabular}{ll}
\toprule
\textbf{Component} & \textbf{Configuration} \\
\midrule
Input probe & NormalSE (256-dimensional) \\
Model dimension \(D\) & 256 \\
Output node-encoding dimension & 256 \\
Polynomial graph filter bank & $K=4$ filters, polynomial degree $d=15$ \\
Conditioning blocks & $B=6$ blocks \\
Local conditioning & 2-layer ResGatedGCN \\
Cross-filter conditioning & 2-layer pairwise-comparison MLP \\
Global conditioning & 2-layer graph-statistics MLP \\
FiLM modulation & Channel-wise scale/shift modulation \\
FiLM residual scale \(s_{\mathrm{res}}\) & Learnable scalar initialized to \(1\) \\
Final fusion & Concatenate $K$ filter branches and project \\
Adjacency operator $A$ & Unnormalized adjacency matrix \\
\bottomrule
\end{tabular}
\end{table}

\subsection{Pretraining Setup}
\label{sec:appendix-pretraining}

CondPSE is pretrained on OGBG-MolPCBA using the same positional/structural pretraining targets as GPSE~\cite{pmlr-v235-canturk24a}. The model reconstructs node-level positional/structural targets and graph-level invariants from NormalSE probes. The node-level targets are ElectrostaticPE, LapPE, RWSE, and HKdiagSE; the graph-level invariants are EigVals and CycleGE. We train the model for 120 epochs using the AdamW optimizer with a batch size of 512, a learning rate of $5 \times 10^{-4}$, weight decay of $1 \times 10^{-4}$, and a cosine learning rate scheduler with warmup.

\subsection{Downstream Adapter and Training Details}
\label{sec:appendix-downstream}

For downstream tasks, the pretrained CondPSE encoder is frozen. The stored CondPSE node encoding is passed through a lightweight adapter and concatenated with projected raw node features before being passed to the downstream backbone.

Across Tables~\ref{tab:expressivity}--\ref{tab:backbone},
baseline values for no encoding, GPSE, LapPE, and RWSE are taken
from the GPSE paper or its public implementation reports, where applicable. CondPSE
results are produced under the corresponding public GPSE evaluation
settings, including dataset splits, seed protocols, and downstream backbone configurations. Backbone architectural hyperparameters, including the number of
layers, hidden dimension, readout/pooling, and backbone-specific
normalization or dropout settings, are inherited unchanged from the corresponding public GPSE configuration files.
During CondPSE evaluation, the learned PSE encoder remains frozen;
only the downstream adapter and downstream backbone are trained.
In learned-PSE settings, the GPSE encoding is replaced by the stored
frozen CondPSE encoding under the corresponding downstream
configuration.

We use the dataset-specific supervised objectives from the
corresponding GPSE downstream configurations: cross-entropy loss
for CSL and EXP, L1 loss for ZINC and PCQM4Mv2-subset, binary
cross-entropy with logits for MolHIV, and masked multi-label
binary cross-entropy with logits for MolPCBA.

For downstream evaluation, CondPSE node encodings are generated
once with a fixed NormalSE probe realization for each graph and
stored. Downstream training and inference reuse these stored
encodings.

For CSL and EXP, downstream models are trained for 200 epochs
using the Adam optimizer with a Reduce-on-Plateau learning rate
scheduler. For molecular datasets, downstream models are trained
using AdamW for 2000, 100, 100, and 100 epochs on ZINC, MolHIV,
MolPCBA, and PCQM4Mv2-subset, respectively, with a cosine decay
learning rate scheduler and warmup. The adapter and training
parameters are reported in Table~\ref{tab:downstream_hparams}.

\begin{table}[H]
\centering
\caption{Downstream adapter and training hyperparameters. Dropout rates are reported before and after the adapter projection. The ZINC row reports the CondPSE adapter/training setting used with the GPS backbone in Table~\ref{tab:downstream}. CondPSE
experiments in the ZINC downstream backbone sweep reuse this
ZINC adapter setting while varying the downstream architecture.}
\label{tab:downstream_hparams}
\scriptsize
\setlength{\tabcolsep}{2.5pt}
\begin{tabular}{llccrcc}
\toprule
\textbf{Dataset} & \textbf{Backbone} & \textbf{Adapter} & $d_{\text{pe}}$ & \textbf{Dropout} & \textbf{Batch Size} & \textbf{Learning Rate} \\
\midrule
CSL & GIN & Linear & 128 & 0.0 / 0.0 & 128 & $2 \times 10^{-3}$ \\
EXP & GIN & Linear & 64 & 0.0 / 0.0 & 128 & $2 \times 10^{-3}$ \\
ZINC & GPS & Linear & 32 & 0.3 / 0.1 & 32 & $1 \times 10^{-3}$ \\
MolHIV & GPS & Linear & 32 & 0.3 / 0.1 & 32 & $1 \times 10^{-4}$ \\
MolPCBA & GPS & 2-layer MLP & 48 & 0.3 / 0.1 & 512 & $5 \times 10^{-4}$ \\
PCQM4Mv2-subset & GPS & Linear & 64 & 0.2 / 0.1 & 256 & $2 \times 10^{-4}$ \\
\bottomrule
\end{tabular}
\end{table}

\subsection*{GenAI Usage Disclosure}

AI tools were used to assist with light editing and grammar. The authors are responsible for all scientific content, experimental results, claims, and final manuscript decisions.

\end{document}